\documentclass{llncs} 
\usepackage{aliascnt} 
\usepackage{algorithm}
\usepackage{algorithmic}
\usepackage{graphicx}
\usepackage{epstopdf}
\usepackage{array}
\usepackage{breakcites}
\usepackage{lipsum}

\usepackage{hyperref}
\makeatletter
\def\UrlAlphabet{%
      \do\a\do\b\do\c\do\d\do\e\do\f\do\g\do\h\do\i\do\j%
      \do\k\do\l\do\m\do\n\do\o\do\p\do\q\do\r\do\s\do\t%
      \do\u\do\v\do\w\do\x\do\y\do\z\do\A\do\B\do\C\do\D%
      \do\E\do\F\do\G\do\H\do\I\do\J\do\K\do\L\do\M\do\N%
      \do\O\do\P\do\Q\do\R\do\S\do\T\do\U\do\V\do\W\do\X%
      \do\Y\do\Z}
\def\UrlDigits{\do\1\do\2\do\3\do\4\do\5\do\6\do\7\do\8\do\9\do\0}
\g@addto@macro{\UrlBreaks}{\UrlOrds}
\g@addto@macro{\UrlBreaks}{\UrlAlphabet}
\g@addto@macro{\UrlBreaks}{\UrlDigits}
\makeatother

\usepackage{threeparttable}

\usepackage[lofdepth,lotdepth]{subfig}

\begin{document}

\title{FLBench: A Benchmark Suite for Federated Learning}

\author{Yuan Liang\inst{1,3,4}, Yange Guo\inst{1,3}, Yanxia Gong\inst{1,3}, Chunjie Luo\inst{2}, Jianfeng Zhan\inst{2} and Yunyou Huang\inst{1,3,}\thanks{Yunyou Huang is the corresponding author}}

\institute{Guangxi Key Lab of Multi-Source Information
Mining \& Security, Department of Computer Science, Guangxi Normal University, Guilin 541004, China. \\
 \email{\{liangyuan,huangyunyou\}@gxnu.edu.cn}, \\
\email{\{gyg2019010376,gyx19980201\}}@stu.gxnu.edu.cn \and
Institute of Computing Technology, Chinese Academy of Sciences,  Beijing, 100190, China \\ 
\email{\{luochunjie, zhanjianfeng\}}@ict.ac.cn \and
BenchCouncil R\&D Lab - Guilin \and 
Guangxi Key Laboratory of Trusted Software, Guilin University of Electronic Technology, Guilin, 541004, China}
%

\maketitle

\begin{abstract}
Federated learning is a new machine learning paradigm. The goal is to build a machine learning model from the data sets distributed on multiple devices--so-called an isolated data island--while keeping their data secure and private. Most existing federated learning benchmarks work manually splits commonly-used public datasets into partitions to simulate real-world isolated data island scenarios. Still, this simulation fails to capture real-world isolated data island's intrinsic characteristics. 
This paper presents a federated learning (FL) benchmark suite named FLBench. 
FLBench contains three domains: medical, financial, and AIoT. By configuring various domains, FLBench is qualified to evaluate federated learning systems and algorithms' essential aspects, like communication, scenario transformation, privacy-preserving, data distribution heterogeneity, and cooperation strategy. Hence, it becomes a promising platform for developing novel federated learning algorithms. Currently, FLBench is open-sourced and in fast-evolution. We package it as an automated deployment tool. 
The benchmark suite is available from \url{https://www.benchcouncil.org/flbench.html}.
\end{abstract}


\section{Introduction}

Google recently proposed the concept of Federated Learning (FL). The main idea is to build a machine learning model from the data sets distributed on multiple devices--so-called an isolated data island--while preventing data leakage~\cite{Konen2016Federated,articlefl,articledeep}. 
FL has become a hot research topic in both industry and academia~\cite{articleqiang,Li2019Federated,2019Advances}. Unfortunately, most existing FL benchmarks work~\cite{Xie2020DBA,Agnostic2019,noniid2019,Bhagoji2018Analyzing,Yurochkin2019Bayesian,FetchSGD2020,Matched2020} manually splits commonly-used public data sets into partitions to simulate isolated data island scenarios~\cite{2020AIBench}; however, they fail to capture the intrinsic characteristics of real-world scenarios. We call an isolated data island scenario a scenario for abbreviation in the rest of this paper.

On the one hand, the statistical data characterization of simulated scenario, which are manually split, are different from those of the real-world one. For example, in centralized training, data can be assumed to be independent and identically distributed (IID). this assumption is unlikely to hold in federated learning settings.
For example, Chandra et al.~\cite{2019SplitFed} use the MNIST and CIFAR-10 datasets to simulate a scenario and assume that the datasets are random, disjoint, and evenly distributed between clients. However, in the real-world scenario, the medical image data will involve the magnetic field intensity issue. For example, the magnetic field intensity at 1.5 T and the magnetic field intensity at 3 T of the same image will show different lesions. The more details are shown in Table~\ref{Tab1} and Table~\ref{Tab1-1}.

Second, without considering the data characterstics, the FL algorithms developed based on the simulated scenario cannot be migrated to a real-world one. For example, in an Alzheimer's diagnosis scenario, it is a CT image--a 3d black and white image, so the FL algorithms developed on the MNIST and CAFAR-10 data sets~\cite{Yurochkin2019Bayesian} are hard to migrate to an Alzheimer's diagnosis.

\begin{table}
\centering
\scriptsize
\caption{The Summary of the Latest FL Publications.}
\begin{threeparttable}
\begin{tabular}{|p{1.2cm}<{\centering}|p{1.8cm}<{\centering}|p{2.0cm}<{\centering}|p{1.0cm}<{\centering}|p{2.5cm}<{\centering}|p{1.5cm}<{\centering}|p{1.2cm}<{\centering}|}
\hline
Venue & Datasets & Simulation approaches &  Isolated data island\tnote{a} & Task & Consistent or not\tnote{b} & real-world scenario\tnote{c} \\
\hline
ICLR~\cite{Xie2020DBA} & MNIST;  CIFAR-10; Tiny-imagenet  & Training: 80\%;   Testing: 20\%; \tnote{d} &  No & Image classification and loan status prediction & No & No \\
\hline
ICML~\cite{Agnostic2019} & Adult dataset, Cornell movie dataset, Penn TreeBank (PTB) dataset, Fashion MNIST & Adult data and Fashion MNIST\tnote{e} & No & Census income forecast, language modeling and image classification  & Yes & Yes \\
\hline
ICLR~\cite{Fair2020} & Public datasets and Synthetic dataset\tnote{f} & $-$ & No & Image classification, emotion analysis, language modeling, vehicle prediction & No & No \\
\hline
ICLR~\cite{noniid2019} & MNIST & MNIST balanced and MNIST un-balanced \tnote{g}& No & Image classification & No & No \\
\hline
\end{tabular}
 
    \end{threeparttable}
\label{Tab1}
\end{table}

\begin{table}[t]
\centering
\scriptsize
\caption{The Summary of Latest FL Publications.(cont.)}
\begin{threeparttable}
\begin{tabular}{|p{1.2cm}<{\centering}|p{1.8cm}<{\centering}|p{2.0cm}<{\centering}|p{1.0cm}<{\centering}|p{2.5cm}<{\centering}|p{1.5cm}<{\centering}|p{1.2cm}<{\centering}|}
\hline
ICLR~\cite{Domain2020} & MNIST, Amazon Review, DomainNet, Office-Caltech10 & 10 Titan-Xp GPU cluster and simulate the federated system on a single machine & No & Image classification, target recognition, DomainNet, emotion analysis & Yes  & Yes \\
\hline 
ICML~\cite{Bhagoji2018Analyzing} & Fashion-MNIST, UCI Adult census dataset & Fashion MNIST and UCI Adult census dataset\tnote{h} &  No & Image classification and census income forecast & No & No \\
\hline
ICML~\cite{Yurochkin2019Bayesian} & MNIST, CAFAR-10 & Randomly divided into J batches\tnote{i} & No & Image classification & No & No \\
\hline
ICML~\cite{FetchSGD2020} & CIFAR-10/100; FEMNIST; PersonaChat & CIFAR and FEMNIST \tnote{j} & No & Image classification, dialogue prediction for personality & No & No \\
\hline
ICML~\cite{Matched2020} & MNIST; CIFAR-10; shakespeare dataset & One centralized node in the distributed cluster is regarded as the data center, and the other nodes are regarded as local clients.\tnote{k} & No & Image classification, language modeling & Image classification: No; Language modeling: Yes & No \\
\hline
\end{tabular}
 
    \end{threeparttable}
\label{Tab1-1}
\end{table}

\begin{tablenotes}
\label{The Footnote of Table1}
        \footnotesize
        \item[a]{\bf{a}}. Data exists in the form of isolated islands in real scenarios.
        \item[b]{\bf{b}}. Whether the features of the simulated scenario are consistent with those of the real one.  
        \item[c]{\bf{c}}. Whether it is used in the real scenario or migrated to the real scenario 
	\item[d]{\bf{d}}. Training: 80\%;   Testing: 20\%;  The training set is equally divided into 100 participants
	\item[e]{\bf{e}}. Adult data: divide the dataset into two domains with and without doctorates; Fashion MNIST: We extract three categories of data subsets: T-shirts, pullovers, and shirts, and then divide this subset into three areas, each containing a garment.
	\item[f]{\bf{f}}. Public datasets: vehicle dataset; text data built from the complete works of William Shakespeare; Omniglot;  tweet data curated from Sentiment 140.
	\item[g]{\bf{g}}. The former is balanced so that the number of samples on each device is the same, while the latter is highly unbalanced. The number of samples between devices follows the power law. 
	\item[h]{\bf{h}}. Fashion MNIST: a 3-layer convolution neural networks (CNN)-based an offline model is used. UCI adult census dataset: uses fully connected neural networks. Set the number of agents K to 10 and 100. When $k = 10$, all agents are selected in each iteration, while when $k = 100$, one-tenth of agents are randomly selected in each iteration.	\item[i]{\bf{i}}. These datasets are randomly divided into J batches. Two partitioning strategies are of interest: (a) uniform partitioning, in which each class in each batch has approximately equal proportions and (b) miscellaneous new partitions with unbalanced batch size and class proportions.
	\item[j]{\bf{j}}.CIFAR: uses 50000 training data points and 10000 validated standard training/test splits. The dataset is divided into 10000 (CIFAR-10) and 50000 (CIFAR-100) clients. Each client has five (CIFAR-10) and one (CIFAR-100) data point from a single target class. In each round, 1\% of the clients participated, resulting in a total batch size of 500 (100 clients of CIFAR-10 have 5 data points, and 500 clients of CIFAR-100 have 1 data point). Federated EMNIST(FEMNIST): classes(upper- and lower-case letters lus digits) which is formed by partitioning the EMNIST dataset such that each client in FEMNIST contains characters written by a single person. 
	\item[k]{\bf{k}}. For the CIFAR-10 dataset, data enhancement (random clipping and flipping) is used, and each image is normalized. We propose two data partitioning strategies  to simulate the joint learning scheme. Homogeneous partitioning: the proportion of each local client is approximately equal in each class. Heterogeneous partitioning: the number of data points and the proportion of classes are unbalanced. For Shakespeare's dataset, we treat each speaking role as a client, resulting in naturally heterogeneous partitions. We preprocess the Shakespeare dataset by filtering out clients with less than 10k data points and sampling a random subset of J = 66 clients. We allocate 80\% of the data for training and merge the rest into the global test set.
      \end{tablenotes}

The common datasets (CIFAR-10, CINIC-10) are partitioned for each participant with statistical methods. Hu et al.~\cite{2020OARF}  shows that the OARF benchmark suite is diverse in data size, distribution, feature distribution, and learning task complexity. Still, it focuses on benchmarking systems instead of algorithms. In addition, Luo et al.~\cite{Luo2019Real} and Hsu et al.~\cite{Classification2020} presume independent and identical distribution, this data type is rare in real life. 

Therefore, this paper calls attention to building an FL benchmark suite to provide various scenarios. 
We propose a benchmark framework with flexible customization and configuration. Covering the three most concerning domains: medical, financial, and AIoT, FLBench provides three scenario benchmarks~\cite{2020AIBench} for developing novel FL systems and algorithms as the real-world scenario is unavailable for most of the researchers. Each scenario benchmark models the critical paths of a real-world application scenario as a permutation of essential modules~\cite{2020AIBench}.  Our suite can evaluate FL system and algorithms' various aspects, including communication, scenario transformation, privacy-preserving, data distribution heterogeneity, and cooperation strategy.  
Table~\ref{Tab2} summarizes the critical differences between FLBench and existing FL libraries and benchmarks. Our key contributions are:

\begin{table}[t]
\scriptsize
\centering
\caption{Comparison with existing federated learning benchmarks.}
\begin{tabular}{|p{1.6cm}<{\centering}|p{1.3cm}<{\centering}|p{1.1cm}<{\centering}|p{1.5cm}<{\centering}|p{1.2cm}<{\centering}|p{1.0cm}<{\centering}|p{0.7cm}<{\centering}|p{1.3cm}<{\centering}|p{1.3cm}<{\centering}|}
\hline
  & Scenario Configuration & Given Fixed Scenario & Customized Scenario & Medicine & Finance & AIoT & Evaluation Metrics & Automated Deployment Tool \\\hline 
FedML~\cite{2020FEDML} & X & $\surd$ & X & X & X & $\surd$ & X & $\surd$  \\\hline 
OARF~\cite{2020OARF} & X &X &X &X &X &X &$\surd$ & $\surd$\\\hline 
IDFL~\cite{Classification2020} & X&X & $\surd$& X& X& $\surd$ &$\surd$ & $\surd$\\\hline  
FVC~\cite{Luo2019Real}  & X &X &X &X &X & $\surd$& X&$\surd$ \\\hline 
FLBench & $\surd$ & $\surd$ & $\surd$ &$\surd$  & $\surd$ & $\surd$ & $\surd$ & $\surd$ \\\hline 
\end{tabular}
\label{Tab2}
\end{table}

1) We propose a configurable FL benchmark suite--FLBench,  covering the three most concerning domains: medical, financial, and AIoT. FLBench can be used to evaluate FL systems and algorithms' different aspects, including communication, scenario transformation, privacy-preserving, data distribution heterogeneity, and cooperation strategy.

2) FLBench provides various customized scenarios for developing novel FL algorithms.

3) FLBench is packaged as an automated deployment tool and can be deployed in mobile, distributed, and standalone manners.

\section{Related Work}

\subsection{Federated Learning}

Federated Learning is to build a machine learning model from the data sets distributed on multiple devices while preventing data leakage~\cite{Konen2016Federated,articlefl,articledeep}. According to data distribution characteristics, FL is mainly divided into horizontal federal learning, vertical federal learning, and federal transfer learning \cite{2020FEDML,articleqiang}. FL benchmarking should consider both systems and algorithms' innovation and performance, so we design a new benchmark suite by evaluating the performance of the FL systems and algorithms from different perspectives. Specifically, we consider the following aspects:

1) {\bf{Communication}}. In the federated network, communication is a key bottleneck. Also, due to the privacy problem of sending original data, the data generated on each device must be kept local. A federated network may consist of many devices, such as millions of smartphones. The speed of communication in the network may be many orders of magnitude slower than local computing. To match the model with the data generated by the devices in the federated network, it is necessary to develop a communication-efficient method to send small messages or model updates iteratively as a part of the training process~\cite{articlefl,chen2019a}.

2) {\bf{Scenario Transformation}}.  Federated learning systems are significantly different from the traditional distributed environment. The main idea of scenario transformation is to transfer a local machine learning model to federated learning settings. Scenario transformation enables people to explore statistical training models on remote devices to suit different scenarios, achieving the purpose of data privacy protection~\cite{bagdasaryan2018how,Domain2020}.

3) {\bf{Privacy-preserving}}. Privacy is often a major concern in federated learning applications. Federated learning takes a step towards protecting the data generated on each device by sharing model updates (such as gradient information) rather than raw data. However, during the whole training process, the model updating communication can still disclose sensitive information to the third party or central server. Although current methods aim to enhance 
federated learning's privacy by using secure multiparty computation or differential privacy, these methods usually provide privacy at the cost of reducing model performance or system efficiency \cite{wang2020federated,li2020practical,triastcyn2019federated,toyoda2019mechanism}. There is a big gap between the theoretical results and the real results.

4) {\bf{Data Distribution Heterogeneity}}. Devices often generate and collect data on the network in a non-IID manner. For example, mobile phone users use different languages in the context of the next word prediction task. Also, the number of data points across devices may vary greatly, and there may be an underlying structure that captures the relationships between devices and their related distributions. This data generation paradigm violates the i.i.d. assumption that is often used in distributed optimization, increases the likelihood of stragglers and may increase the complexity of modeling, analysis, and evaluation~\cite{noniid2019,Matched2020,FetchSGD2020,Domain2020}. Data heterogeneity also includes the characteristic heterogeneity of other data, such as small sample data of intelligent terminal, which cannot form a stable distribution. The characteristic heterogeneity is a 
an issue that has not been considered in FL algorithms benchmarking.

5) {\bf{Cooperation Strategy}}. To fully commercialize federal learning between different organizations, it is necessary to develop a fair platform and cooperation strategy. After establishing the model, the model's performance will be reflected in practical application and recorded in the permanent data recording cooperation strategy, such as blockchain-based ones. The model's effectiveness depends on the organizations' contribution to providing high-quality data; the high-quality model relies upon the federation mechanism distributed to all parties. The high-quality model continues to motivate more organizations to join the data federation~\cite{songprifit2019,wang2019measure,toyoda2019mechanism}, and vice versa.

\subsection{Benchmarks}

In recent years, deep learning and machine learning benchmarks have played an important role in the machine
learning area. Typical benchmarks include AIBench~\cite{gao2018aibench,tang_ISPASS_2021, 2020AIBench,jiang2018hpc,luo2018aiot,hao2018edge},  DAWNBench~\cite{endtoend}, and MLPerf~\cite{MLPerf,MLPerfinproceedings}. These benchmarks have provided various metrics and results for machine learning training and inference. For example, AIBench is a comprehensive AI benchmark suite, distilling real-world application scenarios into AI Scenario~\cite{2020AIBench}, Training~\cite{gao2018aibench,tang_ISPASS_2021}, Inference, and Micro Benchmarks across Datacenter, HPC~\cite{jiang2018hpc}, IoT~\cite{luo2018aiot}, and Edge~\cite{hao2018edge}. AIBench Scenario benchmarks are proxies to industry-scale real-world applications scenarios~\cite{tang_ISPASS_2021}.  However, today's AI faces two major challenges. In most industries, data exists in isolated islands; second, data privacy and security concerns matter.  However, in many cases, we are forbidden to collect, fuse and use data in different AI processing places. How to solve the problem of data fragmentation and isolation legally is a major challenge for AI researchers and practitioners. Therefore, many researchers have proposed a possible solution: safe federated learning~\cite{Konen2016Federated,articlefl,articledeep}. 

We notice that there have been some researches on FL benchmarking, i.e., FedML~\cite{2020FEDML} and OARF~\cite{2020OARF}. Due to the massive gap between simulated and real-world scenarios, it has not been solved well. FedML~\cite{2020FEDML} conducted experiments in different system environments, but they did not detail the benchmarking methodology.  Hu et al. ~\cite{2020OARF} shows their OARF benchmark suite is diverse in data size, distribution, feature distribution, and learning task complexity, but they lack algorithm-level benchmarking. Luo et al.~\cite{Luo2019Real}, and Hsu et al.~\cite{Classification2020} seldom consider essential aspects like communication, scenario transformation, privacy-preserving, data distribution heterogeneity, and cooperation strategy. They mainly focus on the independent and identical distribution of data. Also, they fail to cover federated learning's mainstream scenarios like medical, financial, and AIoT.

\section{FLBench Methodology and Design}

This section presents FLBench methodology, decisions, and implementation.

\subsection{FLBench Methodology}

1) We investigate the most concerning scenarios. The candidate scenarios involve many domains such as medicine and electricity. However, for a benchmark suite, it is impossible and unnecessary to provide all scenarios since they. Besides, providing many scenarios is very costly. Thus, the first step to construct an FL benchmark is selecting several kinds of scenarios to cover FL's different fundamental aspects.

2) According to the output from Step 1), the data generated by several kinds of real-world scenarios need to be collected for constructing the scenarios. Meanwhile, we perform complex data pre-processing, which requires professional domain knowledge, in this step.

3) According to the output from Step 2), we propose configurable  scenarios to evaluate the FL systems and algorithms. For example, the primary concerns about algorithm evaluation are fairness and algorithm robustness. It requires the FL benchmark can provide various scenarios according to the specific researches for every domain. However, it is very costly to construct scenarios for every potential detailed research. Thus, in this step, we designed the user-oriented configurable scenario. 

4) According to the output from Step 3), we construct two main scenarios for evaluating FL algorithms. For an FL benchmark, two main functions are necessary: a)provide a fixed scenario, which refers to a limited set of scenarios provided for all algorithm evaluations in a research direction of an application domain for the fair comparison of FL. b) give an easily-customized scenario for the development of a novel FL algorithm. 
Besides, we propose specific evaluation metrics for every scenario.

5) based on the above outputs, we design and implement an automated deployment tool to deploy the scenario on different platforms.

\begin{figure}[t]
\centering
\includegraphics[width = 1.0\textwidth]{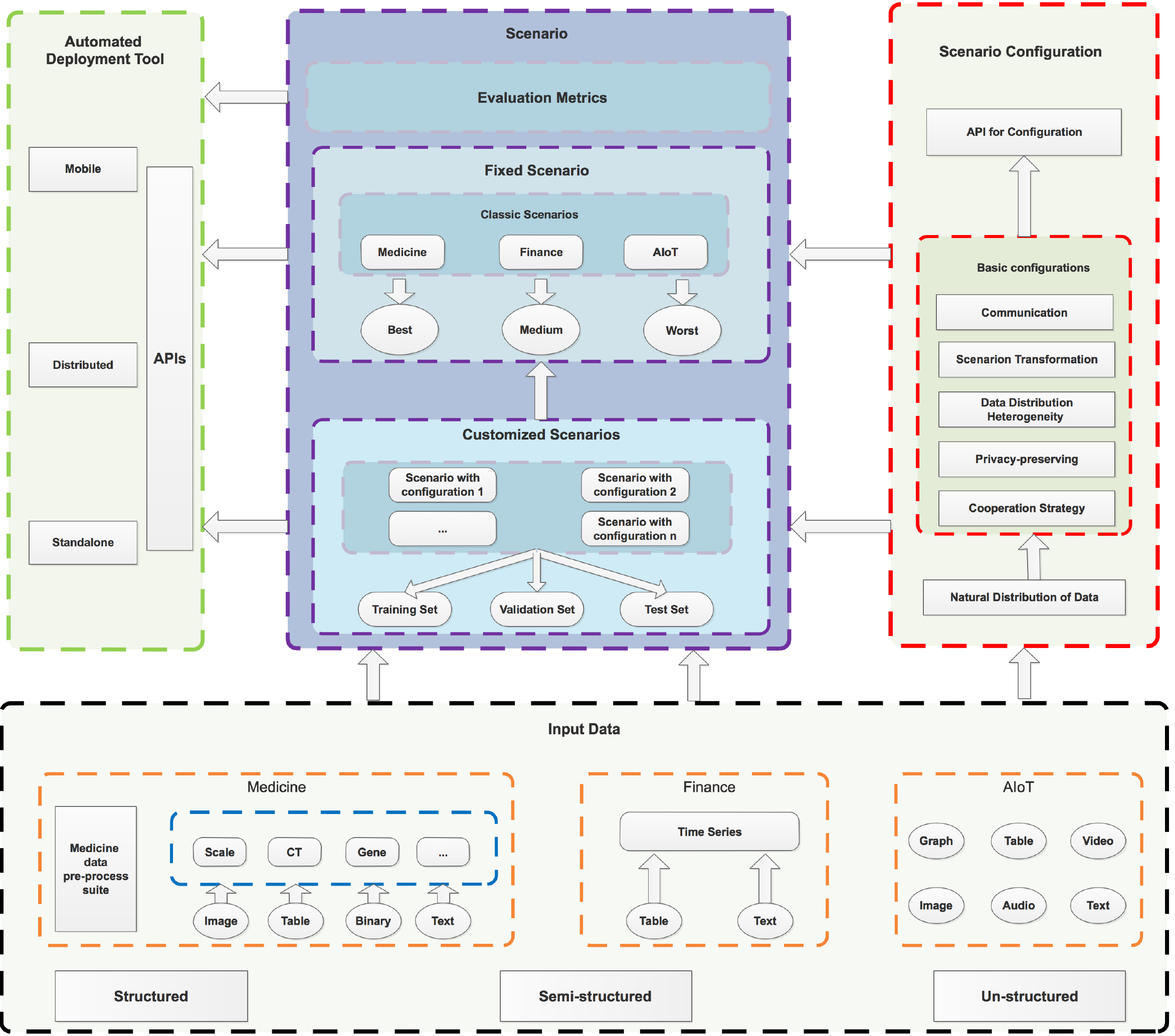}
\caption{The FLBench Framework.}
\label{fig:Framework}
\end{figure}

\subsection{FLBench Design}

Today artificial intelligence still faces two significant challenges. One is that, in most industries, data exists in the form of isolated islands. The other is data privacy and security issue. As analyzed in the FLBench requirements, the FLBench framework (shown in Figure~\ref{fig:Framework}) including four parts.

{\bf{Input Data}}: Most of the current researches on FL are carried out on the simulation scenario, which is constructed by commonly used dataset such as CIFAR-10. 
However, there is a vast difference between the commonly used dataset data and the real-world scenario data in data type and data mode. This considerable difference leads to that FL algorithms developed based on simulated scenarios cannot be migrated to real-world scenarios. We collect data from the three most concerning scenarios to solve this issue, including medicine, finance, and AIoT. Besides, a particular data pre-process tool is necessary for medicine data since medicine data need special processing.

{\bf{Scenario Configuration}}: To achieve the robustness and multi-faceted evaluation of the FL algorithms, we propose a scenario configuration function. First, we analyze the current innovation of FL researches and then classify the innovation directions of FL into the following categories: communication, scenario transformation, privacy-preserving, data distribution heterogeneity, cooperation strategy. Second, for each innovation direction on each domain, we provide a basic configuration according to the natural distribution of data and an API to modify the configuration to simulate various scenarios according to requirements.

{\bf{Automated Deployment Tool}}: We will update FLBench step by step to make it adapt to future development needs. Besides, we continue to expand the benchmark and provide more scenarios and related APIs. We hope that more people will join our benchmark research, which will make our benchmarks suite more perfect and comprehensive.

\subsection{FLBench Implementation}

Currently, FLBench contains: four datasets (medicine: ADNI~\cite{petersen2010alzheimer},MIMIC-III~\cite{johnson2016mimic}; finance: Adult dataset~\cite{mohri2019agnostic}; AIoT: iNaturalist-User-120k~\cite{Classification2020}), one basic configuration file (Alzheimer's diagnosis scenario configuration). The Alzheimer's diagnosis scenario configuration can provide various scenarios for NO-IID (data distribution heterogeneity) research in the medicine domain. 

FLBench is a fully open and evolving benchmark; next, we will provide $3*3=9$ datasets for three domains(medicine, finance, and AIoT), and $3*3*5=45$ basic configuration files on different research aspects, including communication, scenario transformation, privacy-preserving, data distribution heterogeneity, and cooperation strategy. Each configuration file can provide various scenarios according to the requirements of the specific research.

\section{Conclusion}

This paper presents a federated learning benchmark suite named FLBench. 
FLBench contains three domains: medical, financial, and AIoT. By configuring various domains, FLBench is qualified to evaluate federated learning systems and algorithms' essential aspects, like communication, scenario transformation, privacy-preserving, data distribution heterogeneity, and cooperation strategy. We design and implement a configurable benchmark framework, which can deploy on different platforms and provide simple APIs to users. Hence, it becomes a promising platform for developing novel federated learning algorithms. Currently, FLBench is open-sourced and in fast-evolution. We package it as an automated deployment tool. 
The benchmark suite is available from \url{https://www.benchcouncil.org/flbench.html}.




\bibliographystyle{splncs04}
\bibliography{fLbenchmark}

\begin{thebibliography}{10}
\providecommand{\url}[1]{\texttt{#1}}
\providecommand{\urlprefix}{URL }
\providecommand{\doi}[1]{https://doi.org/#1}

\bibitem{bagdasaryan2018how}
Bagdasaryan, E., Veit, A., Hua, Y., Estrin, D., Shmatikov, V.: How to backdoor
  federated learning. In: The 23rd International Conference on Artificial
  Intelligence and Statistics, {AISTATS}. vol.~108, pp. 2938--2948 (2020),
  \url{http://proceedings.mlr.press/v108/bagdasaryan20a.html}

\bibitem{Bhagoji2018Analyzing}
Bhagoji, A.N., Chakraborty, S., Mittal, P., Calo, S.B.: Analyzing federated
  learning through an adversarial lens. In: Proceedings of the 36th
  International Conference on Machine Learning, {ICML}. vol.~97, pp. 634--643
  (2019), \url{http://proceedings.mlr.press/v97/bhagoji19a.html}

\bibitem{chen2019a}
Chen, M., Yang, Z., Saad, W., Yin, C., Poor, H.V., Cui, S.: A joint learning
  and communications framework for federated learning over wireless networks.
  IEEE Transactions on Wireless Communications  \textbf{20}(1),  269--283
  (2021), \url{https://doi.org/10.1109/TWC.2020.3024629}

\bibitem{endtoend}
Coleman, C., Narayanan, D., Kang, D., Zhao, T., Zhang, J., Nardi, L., Bailis,
  P., Olukotun, K., Ré, C., Zaharia, M.: An end-to-end deep learning benchmark
  and competition. Training  \textbf{100} (2017)

\bibitem{gao2018aibench}
Gao, W., Luo, C., Wang, L., Xiong, X., Chen, J., Hao, T., Jiang, Z., Fan, F.,
  Du, M., Huang, Y., Zhang, F., Wen, X., Zheng, C., He, X., Dai, J., Ye, H.,
  Cao, Z., Jia, Z., Zhan, K., Tang, H., Zheng, D., Xie, B., Li, W., Wang, X.,
  Zhan, J.: Aibench: Towards scalable and comprehensive data center {AI}
  benchmarking. In: First BenchCouncil International Symposium on Benchmarking,
  Measuring, and Optimizing (Bench). vol. 11459, pp.~3--9 (2018),
  \url{https://doi.org/10.1007/978-3-030-32813-9\_1}

\bibitem{2020AIBench}
Gao, W., Tang, F., Zhan, J., Wen, X., Wang, L., Cao, Z., Lan, C., Luo, C.,
  Jiang, Z.: Aibench: Scenario-distilling {AI} benchmarking. CoRR
  \textbf{abs/2005.03459} (2020), \url{https://arxiv.org/abs/2005.03459}

\bibitem{hao2018edge}
Hao, T., Huang, Y., Wen, X., Gao, W., Zhang, F., Zheng, C., Wang, L., Ye, H.,
  Hwang, K., Ren, Z., Zhan, J.: Edge aibench: Towards comprehensive end-to-end
  edge computing benchmarking. In: First BenchCouncil International Symposium
  on Benchmarking, Measuring, and Optimizing (Bench). vol. 11459, pp. 23--30
  (2018), \url{https://doi.org/10.1007/978-3-030-32813-9\_3}

\bibitem{2020FEDML}
He, C., Li, S., So, J., Zhang, M., Wang, H., Wang, X., Vepakomma, P., Singh,
  A., Qiu, H., Shen, L., Zhao, P., Kang, Y., Liu, Y., Raskar, R., Yang, Q.,
  Annavaram, M., Avestimehr, S.: Fedml: {A} research library and benchmark for
  federated machine learning. CoRR  \textbf{abs/2007.13518} (2020),
  \url{https://arxiv.org/abs/2007.13518}

\bibitem{Classification2020}
Hsu, T.H., Qi, H., Brown, M.: Federated visual classification with real-world
  data distribution. In: 16th European Conference on Computer Vision - {ECCV}.
  vol. 12355, pp. 76--92 (2020),
  \url{https://doi.org/10.1007/978-3-030-58607-2\_5}

\bibitem{2020OARF}
Hu, S., Li, Y., Liu, X., Li, Q., Wu, Z., He, B.: The {OARF} benchmark suite:
  Characterization and implications for federated learning systems. CoRR
  \textbf{abs/2006.07856} (2020), \url{https://arxiv.org/abs/2006.07856}

\bibitem{jiang2018hpc}
Jiang, Z., Gao, W., Wang, L., Xiong, X., Zhang, Y., Wen, X., Luo, C., Ye, H.,
  Lu, X., Zhang, Y., Feng, S., Li, K., Xu, W., Zhan, J.: {HPC} {AI500:} {A}
  benchmark suite for {HPC} {AI} systems. In: First BenchCouncil International
  Symposium on Benchmarking, Measuring, and Optimizing Symposium(Bench). vol.
  11459, pp. 10--22 (2018), \url{https://doi.org/10.1007/978-3-030-32813-9\_2}

\bibitem{johnson2016mimic}
Johnson, A.E., Pollard, T.J., Shen, L., Li-Wei, H.L., Feng, M., Ghassemi, M.,
  Moody, B., Szolovits, P., Celi, L.A., Mark, R.G.: Mimic-iii, a freely
  accessible critical care database. Scientific data  \textbf{3}(1), ~1--9
  (2016)

\bibitem{2019Advances}
Kairouz, P., McMahan, H.B., Avent, B., Bellet, A., Bennis, M., Bhagoji, A.N.,
  Bonawitz, K., Charles, Z., Cormode, G., Cummings, R., D'Oliveira, R.G.L.,
  Rouayheb, S.E., Evans, D., Gardner, J., Garrett, Z., Gasc{\'{o}}n, A., Ghazi,
  B., Gibbons, P.B., Gruteser, M., Harchaoui, Z., He, C., He, L., Huo, Z.,
  Hutchinson, B., Hsu, J., Jaggi, M., Javidi, T., Joshi, G., Khodak, M.,
  Konecn{\'{y}}, J., Korolova, A., Koushanfar, F., Koyejo, S., Lepoint, T.,
  Liu, Y., Mittal, P., Mohri, M., Nock, R., {\"{O}}zg{\"{u}}r, A., Pagh, R.,
  Raykova, M., Qi, H., Ramage, D., Raskar, R., Song, D., Song, W., Stich, S.U.,
  Sun, Z., Suresh, A.T., Tram{\`{e}}r, F., Vepakomma, P., Wang, J., Xiong, L.,
  Xu, Z., Yang, Q., Yu, F.X., Yu, H., Zhao, S.: Advances and open problems in
  federated learning. CoRR  \textbf{abs/1912.04977} (2019),
  \url{http://arxiv.org/abs/1912.04977}

\bibitem{Konen2016Federated}
Konecn{\'{y}}, J., McMahan, H.B., Ramage, D., Richt{\'{a}}rik, P.: Federated
  optimization: Distributed machine learning for on-device intelligence. CoRR
  \textbf{abs/1610.02527} (2016), \url{http://arxiv.org/abs/1610.02527}

\bibitem{articlefl}
Konecn{\'{y}}, J., McMahan, H.B., Yu, F.X., Richt{\'{a}}rik, P., Suresh, A.T.,
  Bacon, D.: Federated learning: Strategies for improving communication
  efficiency. CoRR  \textbf{abs/1610.05492} (2016),
  \url{http://arxiv.org/abs/1610.05492}

\bibitem{li2020practical}
Li, Q., Wen, Z., He, B.: Practical federated gradient boosting decision trees.
  In: The Thirty-Fourth {AAAI} Conference on Artificial Intelligence, {AAAI}.
  pp. 4642--4649 (2020),
  \url{https://aaai.org/ojs/index.php/AAAI/article/view/5895}

\bibitem{Li2019Federated}
Li, T., Sahu, A.K., Talwalkar, A., Smith, V.: Federated learning: Challenges,
  methods, and future directions. IEEE Signal Processing Magazine
  \textbf{37}(3),  50--60 (2020),
  \url{https://doi.org/10.1109/MSP.2020.2975749}

\bibitem{Fair2020}
Li, T., Sanjabi, M., Beirami, A., Smith, V.: Fair resource allocation in
  federated learning. In: 8th International Conference on Learning
  Representations, {ICLR} (2020),
  \url{https://openreview.net/forum?id=ByexElSYDr}

\bibitem{noniid2019}
Li, X., Huang, K., Yang, W., Wang, S., Zhang, Z.: On the convergence of fedavg
  on non-iid data. In: 8th International Conference on Learning
  Representations, {ICLR} (2020),
  \url{https://openreview.net/forum?id=HJxNAnVtDS}

\bibitem{luo2018aiot}
Luo, C., Zhang, F., Huang, C., Xiong, X., Chen, J., Wang, L., Gao, W., Ye, H.,
  Wu, T., Zhou, R., Zhan, J.: Aiot bench: Towards comprehensive benchmarking
  mobile and embedded device intelligence. In: First BenchCouncil International
  Symposium on Benchmarking, Measuring, and Optimizing (Bench). vol. 11459, pp.
  31--35 (2018), \url{https://doi.org/10.1007/978-3-030-32813-9\_4}

\bibitem{Luo2019Real}
Luo, J., Wu, X., Luo, Y., Huang, A., Huang, Y., Liu, Y., Yang, Q.: Real-world
  image datasets for federated learning. CoRR  \textbf{abs/1910.11089} (2019),
  \url{http://arxiv.org/abs/1910.11089}

\bibitem{MLPerf}
Mattson, P., Cheng, C., Diamos, G.F., Coleman, C., Micikevicius, P., Patterson,
  D.A., Tang, H., Wei, G., Bailis, P., Bittorf, V., Brooks, D., Chen, D.,
  Dutta, D., Gupta, U., Hazelwood, K.M., Hock, A., Huang, X., Kang, D., Kanter,
  D., Kumar, N., Liao, J., Narayanan, D., Oguntebi, T., Pekhimenko, G.,
  Pentecost, L., Reddi, V.J., Robie, T., John, T.S., Wu, C., Xu, L., Young, C.,
  Zaharia, M.: Mlperf training benchmark. In: Proceedings of Machine Learning
  and Systems 2020, MLSys 2020, Austin, TX, USA, March 2-4 (2020),
  \url{https://proceedings.mlsys.org/book/309.pdf}

\bibitem{articledeep}
McMahan, H.B., Moore, E., Ramage, D., y~Arcas, B.A.: Federated learning of deep
  networks using model averaging. CoRR  \textbf{abs/1602.05629} (2016),
  \url{http://arxiv.org/abs/1602.05629}

\bibitem{Agnostic2019}
Mohri, M., Sivek, G., Suresh, A.T.: Agnostic federated learning. In: Chaudhuri,
  K., Salakhutdinov, R. (eds.) Proceedings of the 36th International Conference
  on Machine Learning, {ICML}. vol.~97, pp. 4615--4625 (2019),
  \url{http://proceedings.mlr.press/v97/mohri19a.html}

\bibitem{mohri2019agnostic}
Mohri, M., Sivek, G., Suresh, A.T.: Agnostic federated learning. In: Chaudhuri,
  K., Salakhutdinov, R. (eds.) Proceedings of the 36th International Conference
  on Machine Learning, {ICML}. vol.~97, pp. 4615--4625 (2019),
  \url{http://proceedings.mlr.press/v97/mohri19a.html}

\bibitem{Domain2020}
Peng, X., Huang, Z., Zhu, Y., Saenko, K.: Federated adversaial domain
  adaptation. arXv:1911.02054  (2019)

\bibitem{petersen2010alzheimer}
Petersen, R.C., Aisen, P.S., Beckett, L.A., Donohue, M.C., Weiner, M.W.:
  Alzheimer's disease neuroimaging initiative (adni): clinical
  characterization. Neurology  \textbf{74}(3),  201--209 (2010)

\bibitem{MLPerfinproceedings}
Reddi, V.J., Cheng, C., Kanter, D., Mattson, P., Schmuelling, G., Wu, C.,
  Anderson, B., Breughe, M., Charlebois, M., Chou, W., Chukka, R., Coleman, C.,
  Davis, S., Deng, P., Diamos, G., Duke, J., Fick, D., Gardner, J.S., Hubara,
  I., Idgunji, S., Jablin, T.B., Jiao, J., John, T.S., Kanwar, P., Lee, D.,
  Liao, J., Lokhmotov, A., Massa, F., Meng, P., Micikevicius, P., Osborne, C.,
  Pekhimenko, G., Rajan, A.T.R., Sequeira, D., Sirasao, A., Sun, F., Tang, H.,
  Thomson, M., Wei, F., Wu, E., Xu, L., Yamada, K., Yu, B., Yuan, G., Zhong,
  A., Zhang, P., Zhou, Y.: Mlperf inference benchmark. In: 47th {ACM/IEEE}
  Annual International Symposium on Computer Architecture, {ISCA}. pp. 446--459
  (2020), \url{https://doi.org/10.1109/ISCA45697.2020.00045}

\bibitem{FetchSGD2020}
Rothchild, D., Panda, A., Ullah, E., Ivkin, N., Stoica, I., Braverman, V.,
  Gonzalez, J., Arora, R.: Fetchsgd: Communication-efficient federated learning
  with sketching. In: Proceedings of the 37th International Conference on
  Machine Learning, {ICML}. vol.~119, pp. 8253--8265 (2020),
  \url{http://proceedings.mlr.press/v119/rothchild20a.html}

\bibitem{songprifit2019}
Song, T., Tong, Y., Wei, S.: Profit allocation for federated learning. In: 2019
  {IEEE} International Conference on Big Data (Big Data). pp. 2577--2586
  (2019), \url{https://doi.org/10.1109/BigData47090.2019.9006327}

\bibitem{tang_ISPASS_2021}
Tang, F., Gao, W., Zhan, J., Lan, C., Wen, X., Wang, L., Luo, C., Dai, J., Cao,
  Z., Xiong, X., Jiang, Z., Hao, T., Fan, F., Zhang, F., Huang, Y., Chen, J.,
  Du, M., Ren, R., Zheng, C., Zheng, D., Tang, H., Zhan, K., Wang, B., Kong,
  D., Yu, M., Tan, C., Li, H., Tian, X., Li, Y., Lu, G., Shao, J., Wang, Z.,
  Wang, X., Ye, H.: Aibench: An industry standard {AI} benchmark suite from
  internet services. CoRR  \textbf{abs/2004.14690} (2020),
  \url{https://arxiv.org/abs/2004.14690}

\bibitem{2019SplitFed}
Thapa, C., Chamikara, M.A.P., Camtepe, S.: Splitfed: When federated learning
  meets split learning. CoRR  \textbf{abs/2004.12088} (2020),
  \url{https://arxiv.org/abs/2004.12088}

\bibitem{toyoda2019mechanism}
Toyoda, K., Zhang, A.N.: Mechanism design for an incentive-aware
  blockchain-enabled federated learning platform. In: 2019 {IEEE} International
  Conference on Big Data (Big Data). pp. 395--403 (2019),
  \url{https://doi.org/10.1109/BigData47090.2019.9006344}

\bibitem{triastcyn2019federated}
Triastcyn, A., Faltings, B.: Federated learning with bayesian differential
  privacy. In: 2019 {IEEE} International Conference on Big Data (Big Data). pp.
  2587--2596 (2019), \url{https://doi.org/10.1109/BigData47090.2019.9005465}

\bibitem{wang2019measure}
Wang, G., Dang, C.X., Zhou, Z.: Measure contribution of participants in
  federated learning. In: 2019 {IEEE} International Conference on Big Data (Big
  Data). pp. 2597--2604 (2019),
  \url{https://doi.org/10.1109/BigData47090.2019.9006179}

\bibitem{Matched2020}
Wang, H., Yurochkin, M., Sun, Y., Papailiopoulos, D.S., Khazaeni, Y.: Federated
  learning with matched averaging. In: 8th International Conference on Learning
  Representations, {ICLR} (2020),
  \url{https://openreview.net/forum?id=BkluqlSFDS}

\bibitem{wang2020federated}
Wang, Y., Tong, Y., Shi, D.: Federated latent dirichlet allocation: {A} local
  differential privacy based framework. In: The Thirty-Fourth {AAAI} Conference
  on Artificial Intelligence {AAAI}. pp. 6283--6290 (2020),
  \url{https://aaai.org/ojs/index.php/AAAI/article/view/6096}

\bibitem{Xie2020DBA}
Xie, C., Huang, K., Chen, P., Li, B.: {DBA:} distributed backdoor attacks
  against federated learning. In: 8th International Conference on Learning
  Representations, {ICLR} 2020, Addis Ababa, Ethiopia, April 26-30.
  OpenReview.net (2020), \url{https://openreview.net/forum?id=rkgyS0VFvr}

\bibitem{articleqiang}
Yang, Q., Liu, Y., Chen, T., Tong, Y.: Federated machine learning: Concept and
  applications. ACM Transactions on Intelligent Systems and Technology
  \textbf{10}(2),  12:1--12:19 (2019), \url{https://doi.org/10.1145/3298981}

\bibitem{Yurochkin2019Bayesian}
Yurochkin, M., Agarwal, M., Ghosh, S., Greenewald, K.H., Hoang, T.N., Khazaeni,
  Y.: Bayesian nonparametric federated learning of neural networks. In:
  Proceedings of the 36th International Conference on Machine Learning, {ICML}.
  vol.~97, pp. 7252--7261 (2019),
  \url{http://proceedings.mlr.press/v97/yurochkin19a.html}

\end{thebibliography}

\end{document}